# Analyzing Large language models chatbots: An experimental approach using a probability test

Melise Peruchini

Network and IT department, Federal University of Santa Catarina, melise.peruchini@ufsc.br

Julio M. Teixeira

Graphic Expression department, Federal University of Santa Catarina, julio.teixeira@ufsc.br

This study consists of qualitative empirical research, conducted through exploratory tests with two different Large Language Models (LLMs) chatbots: ChatGPT and Gemini. The methodological procedure involved exploratory tests based on prompts designed with a probability question. The "Linda Problem", widely recognized in cognitive psychology, was used as a basis to create the tests, along with the development of a new problem specifically for this experiment, the "Mary Problem". The object of analysis is the dataset with the outputs provided by each chatbot interaction. The purpose of the analysis is to verify whether the chatbots mainly employ logical reasoning that aligns with probability theory or if they are more frequently affected by the stereotypical textual descriptions in the prompts. The findings provide insights about the approach each chatbot employs in handling logic and textual constructions, suggesting that, while the analyzed chatbots perform satisfactorily on a well-known probabilistic problem, they exhibit significantly lower performance on new tests that require direct application of probabilistic logic.

## 1 INTRODUCTION

The evolution of Large Language Models (LLMs), artificial intelligence systems designed to understand, generate and manipulate human language, have revolutionized the way machines deal with textual prompts and interact with humans. Chatbots based on LLMs, also known as Generative Artificial Intelligence, are currently a notable application of these technologies, and as they become more sophisticated, it becomes increasingly necessary to investigate them. Therefore, new questions arise in attempts to elucidate aspects related to the rapid changes in the ways humans interact with machines.

Simultaneously, in the field of cognitive psychology, studies on how humans process information and make decisions has revealed a variety of biases and heuristics that shape human thinking, notably introduced by the work of Daniel Kahneman and Amos Tversky. Among the fundamental concepts that challenge the notion of human rationality, lies the representativeness heuristic, which describes how people evaluate probabilities, the tendency of people to judge the probability of an event based on how representative it seems, rather than how frequent it actually is. The "Linda Problem" is a thought experiment in which participants assess the probability $P$ of a woman being both a bank teller and an activist in the feminist movement ($X \cap Y$), versus just being a bank teller ($X$). Kahneman and Tversky demonstrated that, contrary to mathematical logic established in the set theory, many participants erroneously consider the option "Linda is a bank teller and activist in the feminist movement" as more likely than just being a bank teller, which is statistically less probable. Thus, the "Linda problem" illustrates how the representativeness heuristics may lead to the phenomena title "conjunction fallacy", which is the failure to recognize that the joint probability of two events cannot exceed the probability of their individual events.

The LLMs' process and generation of text, assessing which sequences of words are more likely or relevant in a given context, encapsulates semantic meanings and contextual relationships, which can be seen as a type of "textual representativeness". Exploring these characteristics may not only lead to a more comprehensive understanding of how LLMs process and generate language but also how they may replicate (or diverge) the human cognitive processes.

This study aims to address the following research question: How do generative artificial intelligence-based chatbots respond to probability questions that typically leads humans to the conjunction fallacy due to the representativeness heuristics? The objective is to conduct an experiment to analyze two LLM chatbots and identify how they handle the given probability test. The analysis methodology consists of examining the dataset of interactions with the chatbots based on three predefined criteria. The purpose of the analysis is to verify whether the chatbots mainly employ logical reasoning that aligns with probability theory or if they are more frequently affected by the stereotypical textual descriptions in the prompts.

Considering that the Linda Problem is well-known and may be present in the chatbots' databases, the "Mary Problem" is formulated, similar to the original, with the same logic but different semantics. The description of this new character allows further exploration on how LLMs chatbots interpret "textual stereotypes" associated with people's descriptions and how this element influences their outputs. However, it should be noted that this research does not intend to delve into judgment biases and the potential effects of stereotype reproduction but rather explores how artificial minds interpret inputs that historically lead humans to illogical conclusions based on the provided context.

The choice for the Linda Problem to evaluate the logical reasoning capacity of chatbots is based on the following reasons: Firstly, this problem is a classic test in cognitive psychology that explores the understanding and application of probability rules. Secondly, set theory is one of the most basic and essential notions for understanding probability. Therefore, using this problem allows for the evaluation of whether the analyzed chatbots possess an adequate understanding of a basic rule of probability.

To address this matter, the following hypotheses are presented:
- Hypothesis 1: The analyzed chatbots predominantly use a probabilistic reasoning approach to solve the given tests, as specified in the prompt, unaffected by the problem semantics;
- Hypothesis 2: The analyzed chatbots exhibit biases and heuristics similar to those observed in humans, prioritizing semantic construction over probabilistic reasoning in their outputs.
- Hypothesis 3: Presenting the same questions to different chatbot models will result in varied responses, reflecting their different capacities of input handling.

The main findings suggests that, while the analyzed chatbots perform satisfactorily on a well-known probabilistic problem, they exhibit significantly lower performance on new tests that require direct application of probabilistic logic. The studied chatbots exhibit a low capacity to make mathematical associations for tests that are not in their training data, showing the inclination for the semantics and textual description over the application of logical reasoning, even when it is stated in the input prompt. This suggests an opportunity to apply novel or alternative training datasets or to implement different training methodologies that can address this type of demand.

## 2 THEORY AND RELATED WORKS

The evolution of Generative Pre-trained Transformer (GPT) architecture and the rapid release of numerous large language models (LLMs), positions these chatbots as technologies with disruptive potential. Transformers use the attention mechanism to assign weights to input data, learning how to create the output based on the most important parts; word prediction direction and contextual information are important factors in word representation methods to language models [24]. Pre-trained models capture the grammatical relationships between words (syntax) and their meanings (semantics) in context, enabling an understanding of both grammar (how words connect and structure) and semantic (how the meanings of words are understood and inferred), resulting in better performance in language processing [24].

The scientific literature extensively explores the biases and stereotypes inherent in data derived from human language [5, 6, 11, 20]. Within recent artificial intelligence research, the literature discusses machine bias with particular focus on ethical issues such as harmful and discriminatory outputs [14, 16, 19]. However, despite the existence of studies analyzing the LLMs from a cognitive heuristic perspective are also discussed in the scientific literature [1, 4, 8, 17], these are comparatively less common, and experiments in this specific context are still insufficient.

Recent studies have demonstrated that LLMs perform well in handling semantic tasks [13, 15] and commonsense reasoning [2] however, some limitations in the reasoning abilities of the LLM chatbot models have also been noted [17], as they are sometimes labeled as "unreliable reasoners" [2] and "stochastic parrots" that do not truly understand the complexity of human language [3]. Studies on the reasoning capabilities of LLMs are still in their early stages but are of growing interest due to the recent popularization of LLMs.

Although there is a perspective in the scientific literature that cognitive heuristics are an intuitive strategy of human thinking that enhances decision accuracy [9, 10], this study's analysis criteria are based on the concept that such cognitive biases lead to illogical interpretations that do not align with probabilistic reasoning [22, 23]. Thus, this study is fundamentally anchored in the concepts of representativeness heuristic [22] and the conjunction fallacy [23], and thus, the widely known "Linda problem" [23] serves as the basis for tests conducted with the chatbots.

As the Linda problem is widely recognized in the cognitive psychology field, some experiments were already conducted with LLMs chatbots [4, 21] providing valuable insights on the subject. However, given the significant advancements in the capabilities of recent chatbot versions compared to previous models and the rapid evolution of these technologies, new experiments are a relevant opportunity to understand how generative AI chatbots handle cognitive biases commonly observed in human decision-making.

## 3 METHODS

This study consists of qualitative empirical research, conducted through exploratory tests with two generative artificial intelligence models, ChatGPT-3.5 from Open AI and Gemini from Google, both in the free version. The object of analysis are the generated outputs. Using a pre-established script, prompts were elaborated to analyze their ability to handle the test called the "Linda problem". The original experiment by Tversky and Kahneman [1983] employed the Linda Problem in two iterations, one with more response alternatives and later, a "short version" with only two options, which makes the question easier to solve. The "extended version" included some simple alternatives as well as those involving conjunctions, and the individuals were asked to rank order the alternatives from the most probable to the least probable after reading the description of the woman called Linda. The mathematical representation of the problem is $P(X \cap Y) \leq P(X)$, and $P(X \cap Y) \leq P(Y)$, where $X$ is the single event that does not fit the stereotype description of the hypothetical person, Y is a single event that fits the stereotype and $X \cap Y$ is the conjunction of the two events. Therefore, option "h" is less probable than options "f" and "c". The architecture of the original problem is presented below:

*Linda is 31 years old, single, outspoken and very bright. She majored in philosophy. As a student, she was deeply concerned with issues of discrimination and social justice, and also participated in anti-nuclear demonstrations.*

*a. Linda is a teacher in elementary school.*
*b. Linda works in a bookstore and takes Yoga classes.*
*c. Linda is active in the feminist movement. (Y)*
*d. Linda is a psychiatric social worker.*
*e. Linda is a member of the League of Women Voters.*
*f. Linda is a bank teller. (X)*
*g. Linda is an insurance salesperson.*
*h. Linda is a bank teller and is active in the feminist movement. (X ∩ Y)*

The detailed description of Linda leads to judging her as more likely to be a feminist than a bank teller, illustrating how descriptions can influence probability judgments. Some of the alternatives provided in the problem are intentionally designed to induce ambiguity. Given the description, it is not possible to determine whether it is more probable that Linda is a psychiatric social worker than an insurance salesperson, for example. These alternatives serve as a distraction. The primary aim was to observe whether participants would commit the conjunction fallacy, rating the option "bank teller and active in the feminist movement" as more probable than simply "bank teller" because it fits better the stereotype of Linda, despite it being statistically less probable. Therefore, in the original experiment, the participants were expected to perceive the conjunction rule among the possible answers from the most critical alternatives [23], which are "c", "f" and "h" in the problem presented above.

Observing that most of the individuals fail to identify the conjunction, and that more options typically make the problem more difficult to solve [12, 23], the short version was elaborated with only two options: a) Linda is a bank teller (X) and b) Linda is a bank teller and is active in the feminist movement (X ∩ Y)

Aware that the Linda problem is well-known and widely recognized, and could, therefore, be included in the datasets on which these models were trained, a new test with similar logic but different textual construction was designed with the intention of masking the question and, consequently, avoiding or mitigating the possibility that the analyzed chatbots retrieved information about the original experiment from their datasets. A second round of experiments was conducted with the "Mary Problem", as presented below:

*Mary is a smart 22-year-old student. She is majoring in natural sciences. She is also a vegetarian, loves jogging and has three dogs.*
*a. Mary is a waitress and is active in the environmental cause. (X ∩ Y)*
*b. Mary is a member of the chess club.*
*c. Mary is active in the environmental cause. (Y)*
*d. Mary plays piano for a hobby.*
*e. Mary is a theater actress.*
*f. Mary is a waitress. (X)*
*g. Mary works in the accounting department.*
*h. Mary travels a lot and likes swimming.*

The critical alternatives in the Mary Problem are: "Mary is a waitress and is active in the environmental cause" (X ∩ Y), and "Mary is a waitress" (X).

The tests were named Linda Problem Short Version (LPSV), Mary Problem Short Version (MPSV), Linda Problem Extended Version (LPEV) and Mary Problem Extended Version (MPEV). The prompts for each test consist of two parts. Firstly, there is a context phrase informing the chatbot that it is a probability test, an attempt to emphasize the expectation of answers with logical reasoning. The prompt input starts as following:

- "I need help with a probability test. Which option is more probable (a. or b.) in the statement below?" (LPSV and MPSV tests)
- "I need help with a probability test. Rank order the following eight descriptions (represented by letters a. to h. below) in terms of the probability that they describe Linda/Mary" (LPEV and MPEV tests).

Secondly, after the context phrase, each prompt includes the actual problem, consisting of descriptions of Linda and Mary, and the alternatives. To ensure the integrity and validity of the results, the alternatives from a) to h) were randomized in each interaction for each test, keeping the content unchanged but altering their order. Randomization eliminates positional bias and ensures that the correct answer is not chosen based on fixed order patterns, but rather on the effective understanding of the question and the content of the options. This increases the level of reliability in validating the chatbot's effectiveness in processing and interpreting the texts, providing a more accurate measure of its response abilities in the provided contexts.

A preliminary experiment was conducted between July 2023 and September 2023 with four randomly chosen chatbots: ChatGPT, Perplexity AI, ChatSonic and Bard (now Gemini). The experiment began with the LPSV and MPSV tests and progressed to the LPEV and MPEV tests, and the preliminary results demonstrated noteworthy performance on the Linda tests, comparatively lower performance in Mary tests and a considerable number of inconclusive answers.

However, for the current study, performed between 04/20/2024 and 04/30/2024, only two models were tested: ChatGPT and Gemini. This decision was influenced both by the similarity between the models (both Language Models with GPT architecture) and by their representativeness as currently noteworthy leading AI technologies. As ChatGPT and Gemini represent the language models from OpenAI and Google respectively, this makes them exemplary subjects to explore the cutting-edge capabilities of conversational AI. Moreover, while both models are built on transformer architectures, they are implemented with distinct optimizations. The different methodologies in data handling and model updating could offer useful insights into their differences and similarities in input handling.

Since the primary object of analysis are the outputs provided by the chatbots, the following sample size formula was used to calculate the number of required interactions ($n$):

$$n = \left(\frac{Z^2\ P\ (1-p)}{E^2}\right)$$

In the formula, $Z$ represents the $Z-value$ corresponding to the confidence level (set at 90%, with a $Z-value$ of approximately 1.645), $E$ represents the margin of error (5%), and $P$ represents the estimated proportion of correct answers (considering a scenario of maximum uncertainty, set at 50%). Using this formula, the sample calculation indicated that approximately 271 interactions are needed for each chatbot. Since the mathematical logic is the same for solving all the tests, meaning $P(X) \geq P(X \cap Y)$, the total number of interactions is divided among the four tests (LPSV, MPSV, LPEV, and MPEV). Thus, the number of unique interactions per test ($n_{test}$) is 68, and the total number of interactions combined for the two chatbots ($n_{total}$) is 542, as presented below:

$$(n_{test}) = \frac{n}{4} = \frac{271}{4} \approx 68$$

$$(n_{total}) = 2 \times n = 2 \times 271 \approx 542$$

However, due to rounding, 544 interactions were conducted between the two chatbots. The dataset containing all interactions is available online (provided in the Data Availability section).

It should be emphasized that no proprietary applications were developed, and thus, it is not possible to examine precisely how or why each chatbot elaborated on such outputs. However, hypotheses and some inferences can be formulated based on the experiment and existing literature about LLMs and Generative AI. It is also worth noting that each test was conducted on different interaction windows to avoid the possibility of the chatbots using the context of the previous conversation to improve their responses.

A metric was established to evaluate the outputs provided by the chatbots, based on three criteria. The first criterion analyzes whether the chatbot ranked the single event ($X$) - "Bank teller" and "Waitress" - as more probable than the conjunction ($X \cap Y$) The second criterion analyzes if the output mentions the conjunction rule or the conjunction fallacy. This provides insights of whether the retrieved data about the conjunction rule is somehow related to or affects their accuracy on the other criteria. The last criterion focuses on analyzing whether the reasoning behind choosing the event ($X$) as more probable is also aligned with probability theory or not.

Each criterion of analysis was evaluated as "Yes", "No" or "Inconclusive". The "Inconclusive" category is used in cases where both options are considered equally probable or in instances of chatbot hallucinations. Table 1 shows an example of the metrics used to analyze each criterion on the tests:

Table 1: Analysis methodology

| Code | Chatbot | Test | The single event (X) is ranked as more probable than the conjunction (X∩Y). | The conjunction rule or the conjunction fallacy is mentioned in the output | The reasoning aligns with the probability theory/conjunction rule |
|---|---|---|---|---|---|
| A0001 | ChatGPT | LPSV | Yes | No | Inconclusive |
| A0002 | ChatGPT | MPSV | No | Yes | No |
| A0003 | Gemini | LPEV | Inconclusive | No | Yes |
| A0004 | Gemini | MPEV | Yes | No | Yes |

Prepared by the authors

The results are described and discussed in Section 4.

## 4 RESULTS AND DISCUSSION

In this section, we present and discuss the results of the exploratory tests (see the linked database to access each interaction with each chatbot).

Since the LPSV test has only two alternatives and is well known, it was expected that the chatbots would be able to solve it easily. In an experiment with students, even those with prior knowledge of statistics and probability, the short version showed a substantial improvement in responses compared to the extended versions, with accuracy increasing from 15% to 64% [12, 23]. The simpler version is easier for humans to answer, and the same applies to the analyzed chatbots. In this experiment, both chatbots mainly answered the LPSV test correctly, selecting option "a) - Linda is a bank teller" as the most probable. The accuracy rate is 98.5% for ChatGPT and 92.6% for Gemini.

The analyzed chatbots frequently referenced the original problem in the LPSV test. This emphasizes the necessity of formulating a different question that conceals the original. Like the LPSV test, MPSV is relatively simple and has only two options as possible answers. However, the constructed text and possible answers are certainly unavailable online or in their datasets, as it was created specifically for this experiment. Therefore, correctly answering the question relies more on applying probability logic rather than making associations based on stored information.

In the MPSV test, the accuracy rate drops to 42.6% for ChatGPT and 58.8% for Gemini, a performance slightly worse than that of humans in the original experiment. It is worth noting that in the MPSV test, Gemini shows a high percent of inconclusive answers, at 32.4%. As mentioned, the outputs are classified as inconclusive in each criterion when the chatbot outputs states that "both options are equally probable" or if there are hallucinations that compromise the answers. Examples of outputs classified as inconclusive in the first criterion are:

*"Option (a) (Linda is a bank teller and is active in the feminist movement) is more probable than option (b). However, it is absolutely possible for Linda to be both a bank teller and a feminist activist. Many people juggle careers with social causes they believe in. This scenario is a classic example of the conjunction fallacy https://en.wikipedia.org/wiki/Conjunction_fallacy. While both options are technically possible, given the additional information, the scenario described in (a) with the feminist activist bank teller is more specific and thus statistically less likely" Gemini Interaction Code A255*

*"Option (a) and (b) are equally probable based on the given information. While Mary's studies (natural sciences) and interests (jogging, vegetarian) suggest a health-conscious and potentially environmentally aware person, this doesn't necessarily translate to her being a waitress. There's also no connection between vegetarianism and the service industry. Therefore, the probability of Mary being a waitress (a) is the same as the probability of her being a waitress involved in environmental causes (b)" Gemini Interaction Code A349*

The overall performance for the first criterion - "($X$) ranked as more probable than the conjunction ($X \cap Y$)" - is presented in Figure 1.

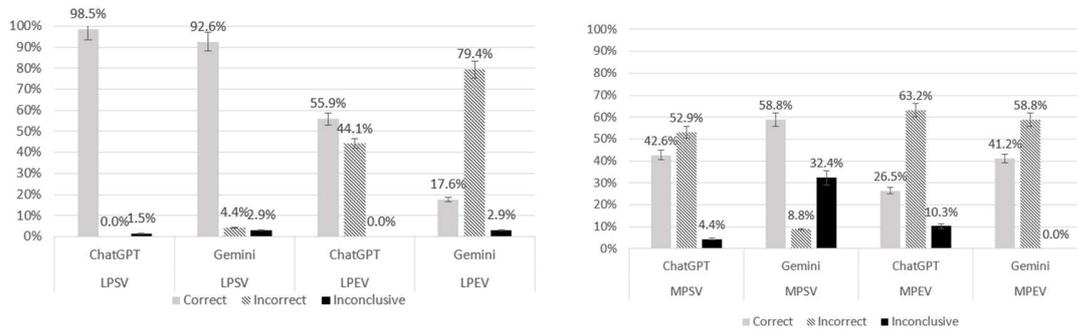

Figure 1: Chatbot accuracy rate on the $P(X \cap Y) \leq P(X)$ criterion. Prepared by the authors.

As shown in Figure 1, the extended versions of the tests significantly lower the accuracy rate, even for the Linda Problem, with 55.9% for ChatGPT and 16.2% for Gemini. The inclusion of an alternative with the event ($Y$) makes the test more difficult to solve because this event is more aligned with the description of the character. For the Mary Problem, Gemini's accuracy rate is 41.2% and ChatGPT's is 25.5%, with a relatively high percentage of inconclusive answers (10.3%)

Regarding the second criterion of analysis, the identification of the conjunction rule or conjunction fallacy, both chatbots performed well in the LPSV tests. However, they are unable to relate Mary's problems to these concepts.

Since the outputs in the LPSV test mention the conjunction rule used to solve the question, one might think that identifying the conjunction in the training data is essential for applying the correct reasoning. However, as observed in the MPSV and MPEV tests, this alone does not influence the ability to choose the single event $X$ as more probable than the conjunction $X \cap Y$. In the MPSV and MPEV tests, the chatbots do not mention the conjunction rule or the conjunction fallacy at all (0%). Examples of correct and incorrect answers despite the conjunction identification are presented below:

*"(...) d. Mary is a waitress - Given no information about her occupation, being a waitress is a relatively common occupation and doesn't contradict any known traits.*
*g. Mary is active in the environmental cause - Mary being a vegetarian suggests she might be environmentally conscious, making this a plausible description.*
*a. Mary is a waitress and is active in the environmental cause - This combine elements from the previous two descriptions and is still plausible"* ChatGPT Interaction Code A443

*"f. Mary is a waitress and is active in the environmental cause. - Possible, but combines the lower probabilities of options a and c."* Gemini Interaction Code A0559

*"Considering the details provided, Linda being a bank teller and active in the feminist movement (option a) seems more likely. This is because the additional detail about her interest in social justice makes activism a more probable fit. This type of problem is a classic example of the conjunction fallacy, where people tend to underestimate the probability of a specific event happening within a broader category"* Gemini Interaction Code A264

*"This problem involves understanding the conjunction fallacy, where people mistakenly judge the conjunction of two events to be more probable than one of the events alone, which defies the laws of probability. (...) 3 - Linda is a bank teller who is active in a feminist movement. (Her activism is more likely than her specific job title.); 4 - Linda is a bank teller. (Given her intelligence and education, this is less likely than her involvement in social or political movements.)" ChatGPT Interaction Code A0368*

The last criterion of analysis consists in observing the reasoning of the outputs, which is considered correct when aligned with the probability rule. For example, choosing *X* before *X ∩ Y* because "*X* is a common job" or "*X* does not conflict with her studies" (as stated in some interactions) is an example of misaligned reasoning. Other examples of correct answer on the first criterion with incorrect reasoning in the same interaction include:

*"Based solely on the information provided, option a - Mary is a waitress - is more probable. If we were to assume that Mary has a part-time job or works to support herself financially, being a waitress might be more common than being active in an environmental cause, based on the information provided" ChatGPT Interaction Code A0167*

*"The more probable option is: b) Mary is a waitress. (...) The statement establishes Mary as a student majoring in natural sciences. This requires dedicated studying and suggests an academic path. Working as a waitress wouldn't necessarily be incompatible with being a student, but it's less likely given her focus on natural sciences. While option (a) adds an element that could connect to her natural science studies (environmental cause), it still includes the less probable element of her being a waitress over a student." Gemini Interaction Code A0107*

From these interactions, it is possible to observe some outputs that illustrates the heavily reliance on the character description in statements such as "Working as a waitress is less likely given her focus on natural sciences", "option (a) still includes the less probable element of her being a waitress over a student" and "if we were to assume that Mary has a part-time job or works to support herself financially, being a waitress might be more common than being active in an environmental cause".

There are also some examples of aligned reasoning, such as "While Mary could be both a waitress and an environmental activist, statistically, the first option is more likely" and "option (b) "Mary is a waitress" appears to be more probable, as it only requires one condition to be true (her being a waitress), while option (a) requires two conditions to be true", as presented in the following interactions:

*"Option (b) 'Mary is a waitress' appears to be more probable, as it only requires one condition to be true (her being a waitress), while option (a) requires two conditions to be true" ChatGPT Interaction Code A0154.*

*"Mary is a waitress and is active in the environmental cause: This combines two possibilities already ranked lower (waitress and environmental cause), making it even less likely" Gemini Interaction Code A0572.*

*"Option (a) "Mary is a waitress" is more probable than option (b). While Mary could be both a waitress and an environmental activist, statistically, the first option is more likely." Gemini Interaction Code A0096.*

The results of the chatbot's accuracy rate (%) across the three criteria from the 544 interactions is synthesized in Table 2 and subsequently discussed.

Table 3: Results of the chatbot accuracy rate across the three criteria

| Test | Chatbot | The single event (X) is ranked as more probable than the conjunction (X∩Y) | The output mentions the conjunction rule or conjunction fallacy | The reasoning aligns with the probability theory/conjunction rule |
|---|---|---|---|---|
| LPSV | ChatGPT | 98.5% | 98.5% | 98.5% |
| LPSV | Gemini | 92.6% | 98.5% | 91.2% |
| MPSV | ChatGPT | 42.6% | 0.0% | 19.1% |
| MPSV | Gemini | 58.8% | 0.0% | 29.4% |
| LPEV | ChatGPT | 55.9% | 26.5% | 54.4% |
| LPEV | Gemini | 16.2% | 42.6% | 20.6% |
| MPEV | ChatGPT | 26.5% | 0.0% | 7.4% |
| MPEV | Gemini | 41.2% | 0.0% | 27.9% |

Prepared by the authors

According to the data in Table 2, ChatGPT correctly solves the MPSV test problem in approximately half of the instances (42.6%); however, these correct answers involve a few incorrect reasoning (19.1%). Similarly, Gemini achieves a 58.8% accuracy rate in the same tests, but in almost half of these cases (29.4%), the reasoning does not adequately align with the probability theory. The results of the reasoning alignment are similar in the MPEV test for both chatbots. From the 26.5% of the times ChatGPT answered the test correctly, only 7.4% of the times the reasoning is also correct. The percentage for Gemini in these criteria is 41.2% and 27.9%, respectively.

In the LPEV test, ChatGPT presents a reasoning aligned with the first criteria in almost the same amount of time (55,9% and 54,4%, respectively), even if not mentioning the conjunction rule in the output at the same rate (26,5%). Gemini, on the other hand, due to the number of inconclusive answers it provided, shows reasoning aligned with probability theory more times that it chooses the correct alternative (20,6% and 16,2%, respectively).

The visual representation for the performance of each chatbot displayed in Table 2 is presented in Figure 2 and Figure 3. Figure 2 provides a granular view of performance across the three key criteria: P(X∩Y) ≤ P(X), Conjunction Identification, and Reasoning. The results are displayed for each chatbot across the distinct tests (LPSV, MPSV, LPEV, and MPEV), allowing for a detailed comparison. Figure 3 complements the view by providing an overall performance for both chatbots across the same tests, highlighting patterns that allows for a more straightforward comparison of the chatbots' abilities.

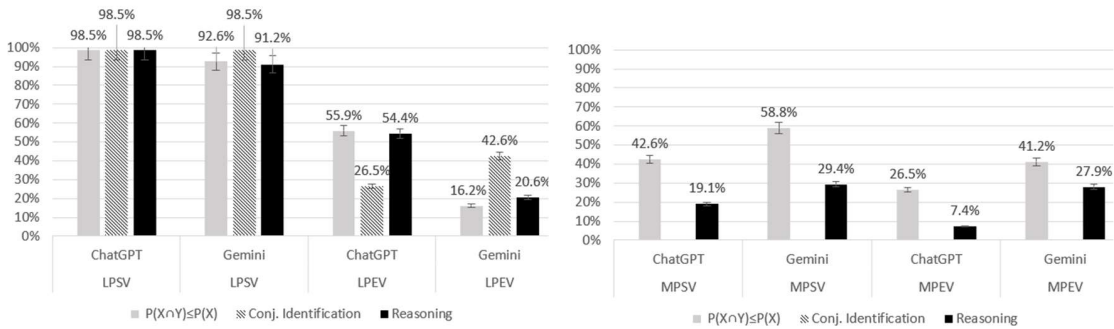

Figure 2: Comparative performance across the three criteria on the "short" versus the "extended" tests. Prepared by the authors.

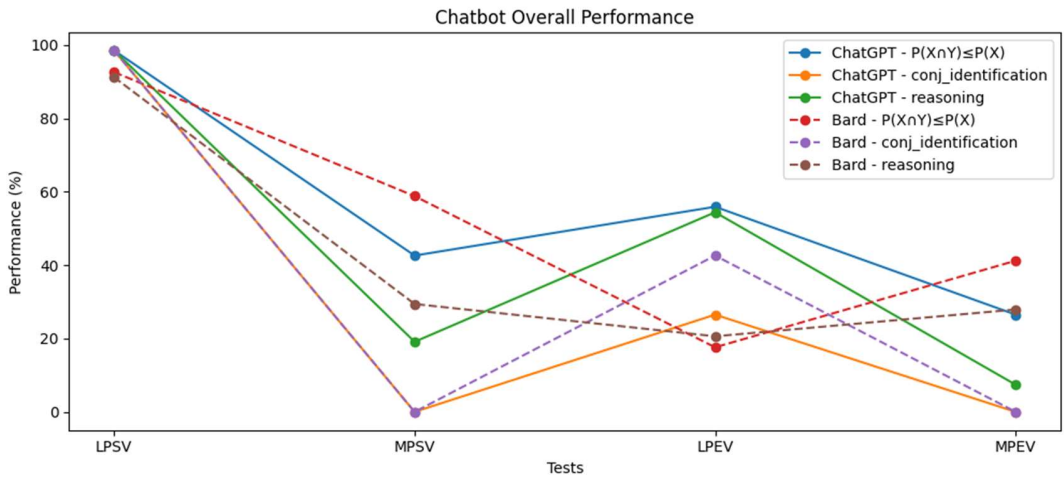

Figure 3: Overall chatbot performance. Prepared by the authors

Mentioning the conjunction fallacy does not necessarily imply avoiding it. The analyzed chatbots' outputs sometimes identify the fallacy but still provide the wrong answer, just as they sometimes give the correct answer without mentioning the conjunction fallacy or the conjunction rule. Similarly, providing the correct answer does not necessarily imply correct reasoning. The variability of the answers provided in the outputs may suggest a certain degree of randomness or non-determinism, which is consistent with existing literature about LLMs [7].

To illustrate the true dependency between answering the question correctly (first criterion) and the correct reasoning (third criterion), three contingency tables were created, and statistical tests were conducted to identify the contingency coefficient (please refer to appendix for the full analysis). The first table considered all tests, the second focused exclusively on the LPVS and LPEV tests, and the third analyzed only the MPVS and MPEV tests. The results showed a significant p-value for each analysis, (respectively: 4.782e-57, 1.2117e-54 and 1.2117e-54) indicating a statistically significant association between the reasoning and answering the question correctly. Specifically, the contingency coefficients were as follows: 0.604 for all tests, 0.674 for the Linda tests, and 0.383 for the Mary tests. While the number of correct answers given by the chatbots significantly depends on the correct reasoning behind the answers in the Linda tests, the strength of this dependency decreases to a moderate level in the Mary tests.

The analyzed data allows for discussion with several implications. Firstly, the analyzed chatbots only identify the conjunction fallacy and representative heuristic in Linda's problems. They are unable to relate Mary's problems to these concepts. An initial experiment with GPT-4, a more robust version, suggests that it can address this issue, as it demonstrates logical reasoning ability even without "stored knowledge" in the database. However, further research is necessary to investigate this more thoroughly. An example of this interaction (please refer to this link to check the full interaction) is:

*"To answer your question (...) we can use the concept of probability known as the conjunction fallacy. In probability theory, the probability of two events happening together (in "conjunction") is always less than or equal to the probability of either one happening alone. That is, P(A and B)≦P(A) and P(A and B)≦P(B). Here, option a is a conjunction of two statements: Mary is a waitress, and she is active in the environmental cause. Option b is just one of these statements: Mary is a waitress. Therefore, option b (Mary is a waitress) is more probable than option a (Mary is a waitress and is active in the environmental cause), because it involves fewer conditions that need to be met, making it more likely in a general sense" ChatGPT4.*

However, the interaction above is not part of the sample analyzed in this study and more interactions are needed to generalize inferences on the subject.

Secondly, the studied chatbots are prone to generating outputs that violate the logic of probability due to the textual composition of the prompt. This is consistent with hypothesis 2 - The analyzed chatbots exhibit biases and heuristics similar to those observed in humans, prioritizing semantic construction over probabilistic reasoning in their outputs - and contrary to hypothesis 1 - The analyzed chatbots predominantly use a probabilistic reasoning approach to solve the given tests, as specified in the prompt, unaffected by the problem semantics. This finding is aligned with previous literature that identify that LLMs mimic human error patterns, prioritizing semantics over logical reasoning [7, 11, 18] and that LLMs appears to present human-like cognitive biases [1, 4].

However, this experiment demonstrates that the chatbot performance on known problems is exceedingly high, therefore, chatbots are capable of learning to provide more appropriate responses and avoid biases if they are effectively trained for it. It is also observed that it is possible to mitigate this "cognitive bias issue", as suggested in the output provided by GPT-4 above. Specific training and fine-tuning may be valuable techniques to optimize the results.

It is also important to note that, according to Shapira et al. [2024], it is imperative to exercise caution when interpreting AI results in tests designed for humans, as these models are built to generate text that appears to be of high quality to human readers. When an LLM performs well on a test originally intended to evaluate cognitive abilities, one cannot automatically infer the same conclusions as for human performance. It is necessary to consider alternative explanations for the model's success, such as the application of different computational methods or shortcuts, which may not reflect a genuine understanding of the test comparable to human cognition [17]

Bender et al [2021] suggests that LLMs are unable to comprehend the complexity and richness of human language, as they depend heavily on statistical patterns to produce sentences. In this research, we assert that although large language models (LLMs) use probabilistic and statistical models to determine sequences of words and phrases, they capture the nuances of human language far more effectively than they handle the given probabilistic problems. This does not imply a statement that the chatbots are able to understand the full complexity of human language; rather, it indicates that there is room for improvement in symbolic aspects of the human language.

In summary, the description of stereotypes existing in the prompt, despite the information that it is a probability test, can impact the responses of chatbots, sometimes tilting them towards the "illogical thinking" akin to the representativeness heuristics observed in humans. The analyzed chatbots often respond based on the stereotyped description of individuals mentioned in the problem, rather than strictly applying logic. Faced with the possibility of stereotype influence, even though some authors consider the human intuition and capability to "classification" - so to speak - as an ability that enhances accuracy, the question of reliability inevitably arises. It arises not so much in relation to the chatbots, which undeniably excel at processing large quantities of text, but regarding the chatbot users, their ability to formulate prompts, understand the functioning of these tools, and to critically analyze the outputs. For example, a prompt that included the mathematical formula might demonstrate a better performance. However, this is not how people typically interact with machines; humans naturally understand semantic language more easily than mathematical symbols. According to Desgupta [2022], reasoning errors in LLMs present a challenge for humans to detect when they reflect our own biases. This research demonstrates that in-depth analysis is necessary to truly understand the underlying reasoning behind the chatbot's performance on problem-solving tasks requiring probabilistic or logical reasoning.

Moreover, this study also illustrates that, despite being effective in optimizing human tasks, more research focusing on experiments with chatbots are essential advances in scientific knowledge to avoid inaccuracies and other potential harmful implications. The findings show that it is not unusual that chatbot models can operate in distinct and sometimes unexpected ways, including cases of hallucinations, and it is important that academics and researchers not only think about the implications of how these models work but also conduct more experiments in the field.

Finally, hypothesis 3 - Presenting the same questions to different chatbot models will result in varied responses, reflecting their different capacities of input handling - proved true in the scope of this experiment. However, this is essentially a qualitative analysis, therefore, further studies with a broader range of chatbot models are needed to support generalization and to provide a more comprehensive understanding of their varying capacities in handling probability questions.

The findings of the experiment present an opportunity to advance in more context-specific development and training of chatbot models. These findings can also be useful for chatbot users as they provide a deeper understanding of how the models operate, and which context or interaction format is most suitable, allowing users to make informed decisions that meet their interaction needs. We emphasize the importance of understanding AI functioning to make informed decisions, which is essential not only to avoid inaccuracies but also to prevent potentially harmful outcomes.

## 5 FINAL CONSIDERATIONS

The main findings of this study reveal that while the analyzed chatbots, ChatGPT and Gemini, perform satisfactorily on well-known probabilistic problems such as the Linda Problem, they exhibit significantly lower performance on new tests like the Mary Problem, in contexts requiring direct use of probabilistic logic. A deeper analysis of the reasoning behind the responses reinforces their semantic preference over logical reasoning. This is especially relevant because this kind of evaluation heavily depends on the user's knowledge about the generated output. The value of this study also lies in demonstrating that a superficial analysis using well-known problems may present an inaccurate assessment of chatbot performance. Similar and future studies that use recognized problems in chatbot testing should take these particularities into account to avoid imprecise results.

For users, these findings demonstrate that while chatbots can be useful for general and well-known problems, caution is required when relying on them for new or complex problems, especially with less sophisticated prompts. Users should be aware that the quality of the chatbot's reasoning can vary significantly and that a deeper understanding and critical evaluation of the chatbot's responses are essential to ensure reliable outcomes.

The findings are also important for chatbot developers, since it presents a critical area for improvement in the development of generative AI. A better understanding about the aspects that affect LLMs outputs can provide useful insights to developers. To enhance the performance of the chatbots, we suggest incorporating more problems of probability, statistics, and mathematical logic into the training data, ensuring that it includes a wide variety of problems that require the direct application of probabilistic and statistical rules rather than just semantic interpretation. Additionally, there is an opportunity for the development of specific algorithms for recognizing and applying mathematical rules and other statistical principles. Moreover, improving the chatbots' ability to disambiguate between problems requiring a semantic approach and those requiring a mathematical approach may be relevant for more accurate and relevant responses.

This study has some limitations that can be further explored in future research. Firstly, it is important to note that it represents a specific snapshot at a particular moment, and its value is tied to the conditions present during the research. The complexity of how these chatbots process text and their continuous adaptation capabilities may lead to significant variations in results over time. Hence, temporal limitations and the dynamic nature of the technologies analyzed are factors that could influence outcomes in future studies. Secondly, to advance our understanding of these technologies, future studies may expand the repertoire of tests or validate the same tests in other chatbots, along with specific techniques such as fine-tuning and Chain-of-thoughts.

This work contributes to the body of research on the performance of Large Language Models / Generative AI by providing an analysis of problem-solving in logic and probability. The experiments with exploratory tests help understand how LLMs handle logic and textual constructions. We encourage the academic and professional community to make efforts in similar experiments, aiming for the replication and expansion of this study by diversifying contexts and continuously evaluating these technologies. It is hoped that this study serves as an encouragement for continuous testing and experimentation, and the advance of scientific knowledge on the field.

## ACKNOWLEDGMENTS

This study was financed in part by the Coordenação de Aperfeiçoamento de Pessoal de Nível Superior - Brasil (CAPES) - Finance Code 001.

# A APPENDICES

## A.1 Dataset

1. Dataset before probabilistic imputation (raw)
2. Dataset after probabilistic imputation for statistical analysis

## A.2 Convergence tables and statistical analysis

Table 1

| All the tests | Z: Yes | Z: No | Total |
|---|---|---|---|
| X: Yes | 232 | 92 | **324** |
| X: No | 5 | 215 | **220** |
| **Total** | **237** | **307** | **544** |

**Total**
count 3.000000
mean 362.666667
std 165.424706
min 220.000000
25% 272.000000
50% 324.000000
75% 434.000000
max 544.000000
P-value: 4.7823571408875834e-57
Contingency coefficient: 0.563699940757736

Table 2

| LPSV and LPEV | Z: Yes | Z: No | Total |
|---|---|---|---|
| X: Yes | 194 | 5 | **199** |
| X: No | 5 | 87 | **92** |
| **Total** | **199** | **92** | **291** |

**Total**
count 3.000000
mean 194.000000
std 99.594177
min 92.000000
25% 145.500000
50% 199.000000

75% 245.000000

max 291.000000

P-value: 1.2116970093206095e-54

Contingency coefficient: 0.6740823153892881

Table 3

| **MPSV and MPEV** | Z: Yes | Z: No | Total |
|---|---|---|---|
| X: Yes | 38 | 87 | **125** |
| X: No | 0 | 128 | **128** |
| **Total** | **38** | **215** | **253** |

**Total**

count 3.000000

mean 168.666667

std 73.050211

min 125.000000

25% 126.500000

50% 128.000000

75% 190.500000

max 253.000000

P-value: 4.374480695583248e-11

Contingency coefficient: 0.3827991863528407